\documentclass[10pt,twocolumn,letterpaper]{article}

\usepackage[pagenumbers]{cvpr} 

\usepackage[dvipsnames]{xcolor}

\definecolor{cvprblue}{rgb}{0.21,0.49,0.74}
\definecolor{myred}{rgb}{0.96,0.,0.09}
\definecolor{myyellow}{rgb}{1.0,0.81,0}
\definecolor{myblue}{rgb}{0.196,0.086,0.690}
\definecolor{mygreen}{rgb}{0.145,0.835,0.0}

\usepackage[pagebackref,breaklinks,colorlinks,urlcolor=cvprblue,linkcolor=cvprblue,citecolor=cvprblue]{hyperref}
\usepackage{dashrule}
\usepackage{tikz}
\usepackage{multirow}
\usepackage{pifont}
\usepackage{bbding}
\usepackage{makecell}
\usepackage{nicematrix}
\newcommand{\mf}[1]{\mathbf{#1}}
\newcommand{\mc}[1]{\mathcal{#1}}
\newcommand{\mb}[1]{\mathbb{#1}}

\def\paperID{****} 
\def\confName{CVPR}
\def\confYear{2024}

\title{Weakly Supervised Video Individual Counting}

\author{
 \textit{Xinyan Liu$^{1}$ \quad Guorong Li$^{1}$\thanks{Corresponding author.} \quad Yuankai Qi$^{2}$} \quad  Ziheng Yan$^{1}$\\\quad Zhenjun Han$^{1}$ \quad Anton van den Hengel$^{2}$ \quad Ming-Hsuan Yang$^{5}$ \quad Qingming Huang$^{1,3,4}$\\
$^{1}$ \small University of Chinese Academy of Science, Beijing, China \\
$^{2}$\small Australian Institute for Machine Learning, The University of Adelaide \\
$^{3}$\small Key Lab of Intell. Info. Process., Inst. of Comput. Tech., CAS, Beijing, China \\
$^{4}$\small Peng Cheng Laboratory, Shenzhen, China, $^{5}$University of California, Merced \\
{\small \{liuxinyan19,yanziheng21\}@mails.ucas.ac.cn, \{liguorong, hanzhj, qmhuang\}@ucas.ac.cn, qykshr@gmail.com,} \\
\small {Anton.vandenHengel@adelaide.edu.au, mhyang@ucmerced.edu}
}
\begin{document}
\maketitle
\begin{abstract}
Video Individual Counting (VIC) aims to predict the number of unique individuals in a single video.
Existing methods learn representations based on trajectory labels for individuals, which are annotation-expensive. 
To provide a more realistic reflection of the underlying practical challenge, we introduce a weakly supervised VIC task, wherein trajectory labels are not provided. Instead, two types of labels are provided to indicate traffic entering the field of view (inflow) and leaving the field view (outflow).
We also propose the first solution as a baseline that formulates the task as a weakly supervised contrastive learning problem under group-level matching. In doing so, we devise an end-to-end trainable soft contrastive loss to drive the network to distinguish inflow, outflow, and the remaining.
To facilitate future study in this direction, we generate annotations from the existing VIC datasets SenseCrowd and CroHD and also build a new dataset, UAVVIC.
Extensive results show that our baseline weakly supervised method outperforms supervised methods, and thus, little information is lost in the transition to the more practically relevant weakly supervised task.
The code and trained model will be public at \href{https://github.com/streamer-AP/CGNet}{CGNet}
\end{abstract}
    
\section{Introduction}
\label{sec:intro}

Video Crowd Counting (VCC) has garnered much interest due to its broad range of practical applications, particularly in crowd safety management. 
This task requires a model to count the number of people in each frame of a video. A limitation of VCC~\cite{TAN,LSTN,ConvLSTM} is that it gives an imprecise estimate of the number of unique individuals appearing in a video sequence, as people are counted multiple times if they appear in several frames. 
To overcome this drawback, Video Individual Counting (VIC) was introduced, wherein a method must count the total number of people with unique identities appearing in a video sequence.

\begin{figure}[!bt]
     \centering
     \begin{subfigure}[b]{0.22\textwidth}
         \centering
         \includegraphics[width=\textwidth]{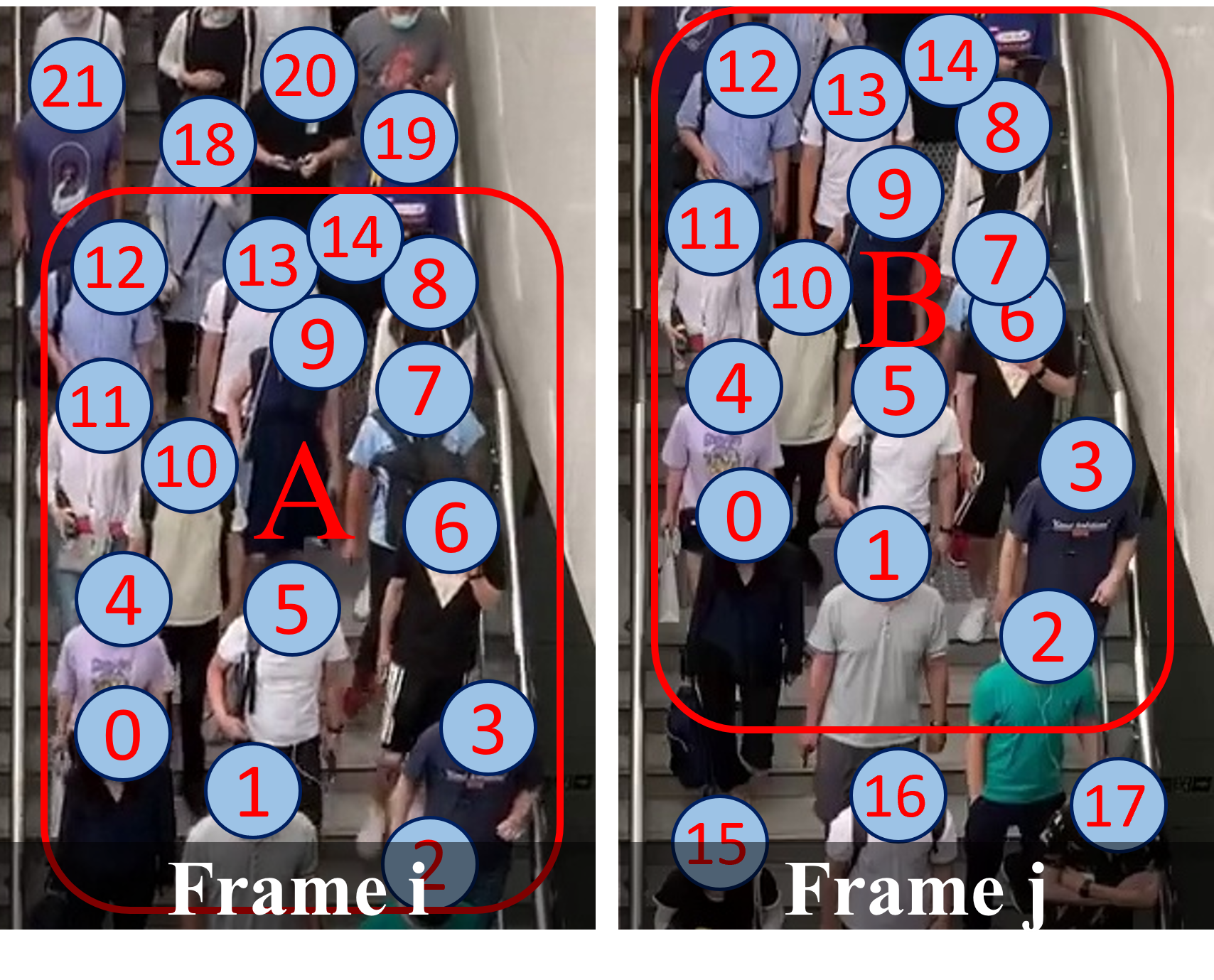}
         \caption{Annotation of VIC method}
         \label{fig:1a}
     \end{subfigure}
     \hfill 
     \begin{tikzpicture}
       \draw[dotted, thick] (0,0) -- (0,3.6); 
     \end{tikzpicture}
     \hfill
     \begin{subfigure}[b]{0.22\textwidth}
         \centering
         \includegraphics[width=\textwidth]{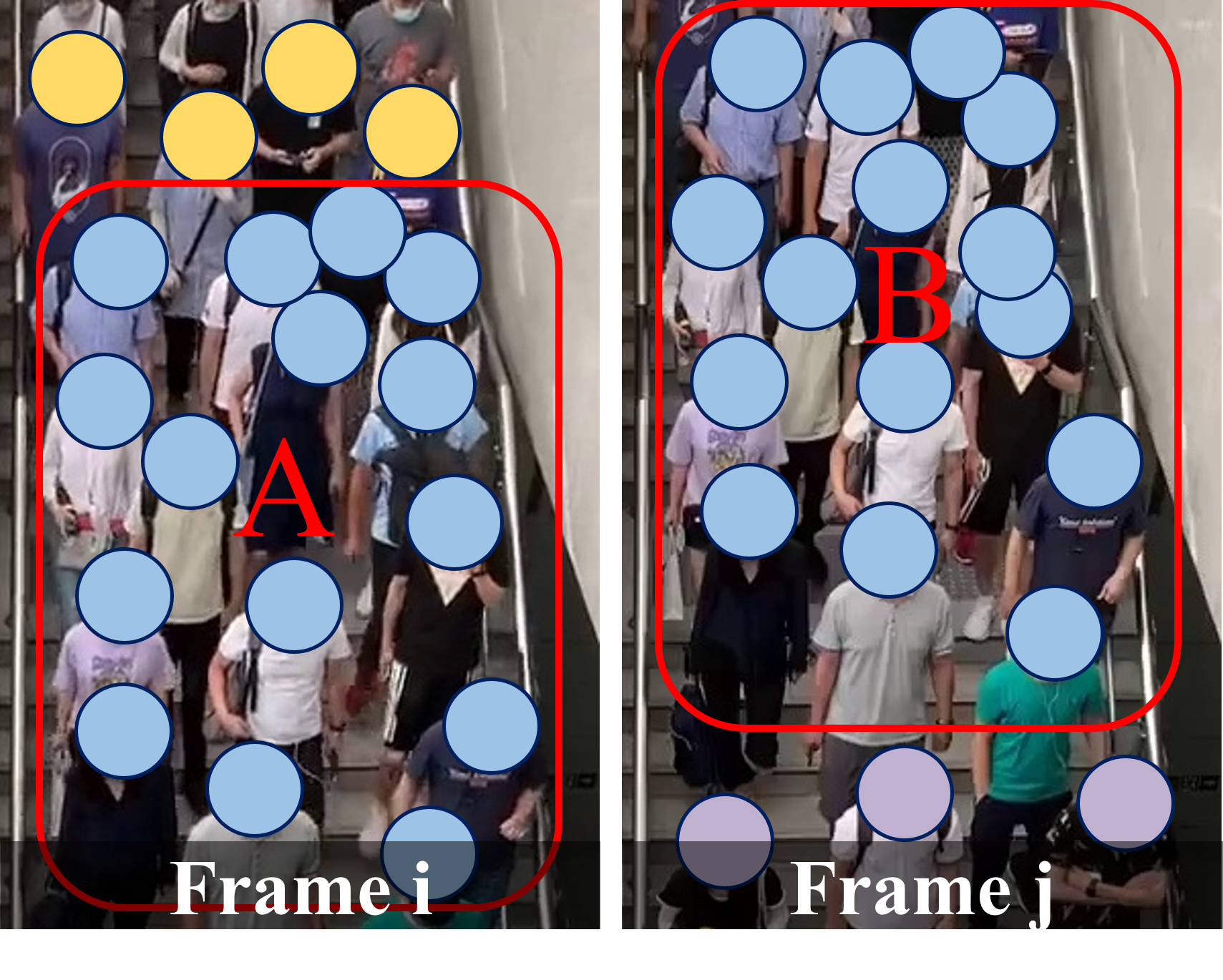}
         \caption{Annotation of our method}
         \label{fig:1b}
     \end{subfigure}
    \caption{Existing VIC method requires a unique label indicating the position of each human in each frame, and these labels are consistent across frames. Weakly supervised VIC only requires labels indicating each human position and whether they are an inflow/outflow pedestrian (purple/yellow in (b)) in the current frame. The transition from individual-level labels to identity-agnostic group-level labels represents a significant reduction in the labeling effort.}
    \label{fig:annotation}
    \vspace{-5mm}
\end{figure}
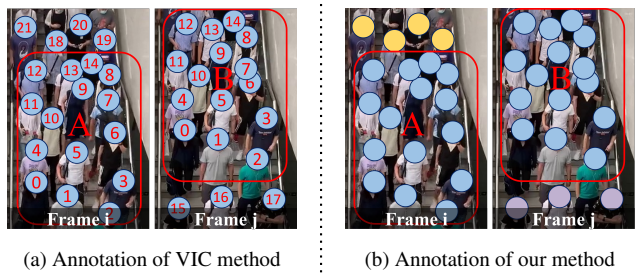

The apparent approach to crowd counting is to count the people who appear in the first frame and add the number of people who come into the camera's field of view in later frames (the inflow). 
Following this principle, Han \etal~\cite{DRVIC} devised DRNet, the only existing published method for VIC, identifying repeated observations of individuals across consecutive frames based on their appearance, simultaneously predicting inflow and outflow. Although the number of outflows does not contribute to the final count, DRNet finds it helps determine individual associations.
%
Han \etal~\cite{DRVIC} 
also explored using conventional multi-object tracking~(MOT) methods~\cite{Tracking_Pedestrian_Heads,High-Speed_tracking-by-detection,CRFtracking,fairmot,phdtt} to tackle the VIC problem. Unfortunately, it turns out MOT methods suffer from poor accuracy. 
More notably, DRNet and MOT-based methods require trajectory labels~(or similar individual association labels) to supervise identity association, which is highly annotation-expensive.

Our key insight is that counting people in the previous and current frames does not require accurate identity associations for those appearing in both frames.
For example, as long as we can predict individual 15$\sim$17 as entering in \cref{fig:1a}, even if we arbitrarily associate individuals 0$\sim$14 in frame $i$ to individuals 0$\sim$14 in frame $i$, we can still correctly infer the crowd counts.
{\em We can still count crowds if we relax individual-level identity associations to group-level associations and do not require trajectory annotations.}
Thus, we just need to annotate inflow and outflow people (then we can derive the individual exists in two frames), which reduces annotation costs compared to creating individual pairwise target associations for each observed pedestrian between neighboring frames. We name crowd counting with such annotations as Weakly supervised Video Individual Counting~(WVIC).
%

To address the WVIC task, we propose a benchmark method based on Contrastive learning with Group-level matching, namely CGNet. 
%
%
We design a soft contrastive loss to drive the network to learn discriminative representations that can facilitate identifying the required group associations and, thus, the inflow. 
Moreover, to better represent each individual, we design a memory-based individual count predictor, where historical templates of individuals are stored and updated in memory to enhance the robustness of association during inference. 
Considering that the two existing datasets for VIC are all captured by static cameras, we collect a weakly supervised VIC dataset based on moving UAVs, named UAVVIC. 
This dataset provides about 400,000 inflow/outflow and bounding box labels for four categories: pedestrian, car, bus, and van. Although devised for crowd counting, our proposed baseline method also performs well in counting other objects.

\vspace{1mm}
Our main contributions are as follows:
\begin{itemize}[leftmargin=2pt,itemindent=0.7cm]
    \item We propose WVIC, a weakly supervised video individual counting task. This task does not require expensive per-target trajectory annotations and only requires two types of identity-agnostic annotations.
    
    \item We automatically reannotate two existing datasets, CroHD and SenseCrowd, and collect a new dataset, UAVVIC, to pave the way for future studies.
    
    \item We propose a strong baseline, CGNet, equipped with a newly designed group level matching soft contrastive loss, performing favorably against the supervised methods on the three datasets mentioned above. 
\end{itemize}
\section{Related Works}
\label{sec:related}

\noindent\textbf{Video Crowd Counting}~(VCC) estimates the number of people in each video frame. Most of the existing VCC methods~\cite{LSTN, TAN, ConvLSTM,clrnet,zheng2018cross,Counting_People_CrossingLine,early_cross_line_work1,early_cross_line_work2,early_cross_line_work3} can be divided into two categories according to their solved problems: region of interest~(ROI) and line of interest~(LOI).
ROI methods~\cite{LSTN, TAN, ConvLSTM,clrnet} detect pedestrians within a specific region~(or the whole image), and they focus on leveraging temporal context information to improve the prediction of the current frame. 
LSTN~\cite{LSTN} models the group flow of crowds in local regions. 
Xiong \etal in \cite{ConvLSTM} fuse history frame features using a ConvLSTM. TAN~\cite{TAN} explores context information from adjacent density maps. CLRNet proposes a local self-attention module to help the model focus on highly related regions in adjacent frames. ROI methods can not be used in the VIC task.
LOI methods ~\cite{zhao2016crossing,zheng2018cross,Counting_People_CrossingLine,early_cross_line_work1,early_cross_line_work2,early_cross_line_work3} counts people passing through a specific line. Most of the existing LOI methods~\cite{early_cross_line_work1,early_cross_line_work2,early_cross_line_work3} apply blobs to crowds, counting objects when the whole blobs cross the line.
Ma \etal in \cite{Counting_People_CrossingLine} sample fixed line width areas within temporal image slices and accumulated crowds in this area. 
Zheng \etal \cite{zheng2018cross} instead sum the number of people within neighboring blocks near a line based on local velocity. Zhao \etal in \cite{zhao2016crossing} train a crowd velocity map predictor with the help of trajectory labels. In practice, LOI methods perform poorly on the VIC task as people can enter or leave the field of view in any direction, making it difficult to find a suitable virtual line.

\noindent\textbf{Multiple Object Tracking} (MOT) aims to predict the trajectories for multiple targets in a video. 
%
Object association is a main challenge in MOT, establishing correspondences between objects across frames to generate targets' trajectories. 
Existing object association methods in MOT include Probabilistic Data Association (PDA)~\cite{PDA1, PDA2}, Joint Probabilistic Data Association (JPDA)~\cite{JPDA1, JPDA2, JPDA3, JPDA4}, and graph matching methods like the Hungarian algorithm~\cite{Hungarian1, Hungarian2}.
Recent research has started to use more powerful feature representations to enhance the robustness of data association. Deep learning-based methods like LSTM can model temporal information to handle long-term occlusion~\cite{LSTM1,LSTM2, LSTM3}. 
Many works~\cite{tracklet1, tracklet2, tracklet3} also focus on the association of tracklets. Specifically, they associate multiple tracklets over longer periods using a graph model to combine the final trajectory, which can relieve the effects of lost tracked targets. 
These methods perform poorly on the VIC task, however, due to the inevitable ID switching problem~\cite{DRVIC}.
Additionally, they require trajectory labels to train their models.

\noindent\textbf{Video Individual Counting}~(VIC). To the best of our knowledge, DRNet~\cite{DRVIC} is the only method for the VIC task. 
It predicts the number of initial pedestrians and matches pedestrians in adjacent frames by adapting a differentiable optimal transport loss. Although the VIC task only requires the total count in the video,  DRNet needs individual matching between frames. 
WVIC avoids the requirement for such matching annotation, which enables easy expansion of dataset scales and thus facilitates future large deep learning models.

\begin{figure*}[!tb]
    \centering
    \includegraphics[width=0.9\textwidth]{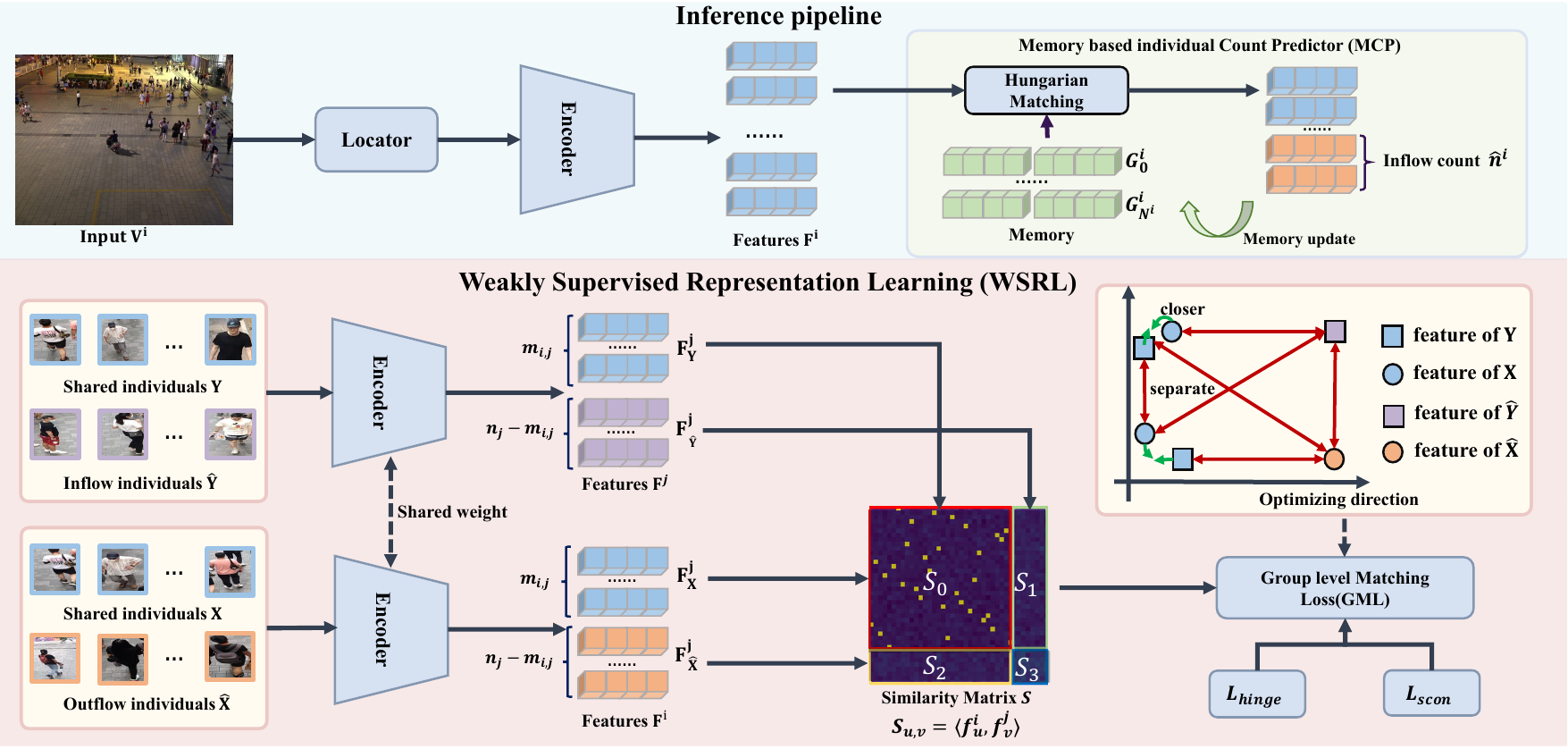}
    \caption{The inference pipeline of our CGNet and the weakly supervised representation learning method~(WSRL).
    The pipeline comprises a frame-level crowd locator, an encoder, and a Memory-based individual count predictor~(MCP). The locator predicts the coordinates for pedestrians. The encoder generates representations for each individual, and MCP predicts inflow counts and updates the individual templates stored in the memory.
    To pull the matched groups ($\mathbf{X}$ and $\mathbf{Y}$) closer and push away individual pairs from unmatched groups, 
    WSRL exploits inflow and outflow labels to optimize the encoder with a novel Group level Matching Loss~(GML), which consists of a soft contrastive loss~($\mc{L}_{scon}$) and a hinge loss~($\mc{L}_{hinge}$).
    }
    \label{fig:pipeline}
     \vspace{-5mm}
\end{figure*}

\section{Task Formulation of WVIC}

\label{sec:problem}
Given a video $\mf{V}$ consisting of $T$ frames ${ \mf{V}^1, \cdots, \mf{V}^T }$, the label information comprises coordinate/inflow/outflow data for each individual.
%
Specifically, letting $\mf{P}^i=[ p^i_1,..., p^i_{n_i} ]$ denote the annotation of the coordinates of the center of the people in frame $\mf{V}^{i}$, we sample the frame pairs $\{(\mf{V}^{1}, \mf{V}^{\delta+1}), (\mf{V}^{\delta+1},\mf{V}^{2\delta+1}),...,(\mf{V}^{T-\delta},\mf{V}^{T})\}$, where $\delta$ is an integer frame offset. The weakly supervised inflow label  $\mf{I}^{i}=[ \mf{I}^{i}_1,...,\mf{I}^i_{n_i} ]$ and outflow label $\mf{O}^{i}=[ \mf{O}^{i}_1,...,\mf{O}^{i}_{n_i}]$. 
If the $u$-th person in frame $i$ 
doesn't appear in $\mf{V}^{i-\delta}$, then $\mf{I}_u^i=1$; otherwise, $\mf{I}_u^i=0$. For $\mf{O}^{i}$, if the $u$-th person in frame $V^i$ doesn't appear in $\mf{V}^{i+\delta}$, then $\mf{O}_u^i=1$; otherwise $\mf{O}_u^i=0$.
If, by a slight abuse of notation, we refer to people by their coordinates in particular images, we see that the set $\{p_u^i| \mf{O}_u^i=0\}$ and $\{p_u^{i+\delta}| \mf{I}_u^i=0\}$ represent the people that appear both in $\mf{V}^{i}$ and $\mf{V}^{i+\delta}$. 
The goal of Weakly-supervised Video Individual Counting~(WVIC) is to predict the number of unique individuals ($\hat{N}$) in a video provided a series of its frames with $\delta$ as the sampling rate. 

\section{A Benchmark Method}
\label{sec:method}
As shown in \cref{fig:pipeline}, the inference pipeline of our CGNet consists of three components: an image-level locator to generate coordinates for pedestrians, an encoder to produce representations based on pedestrians' coordinates, and a Memory-based individual Count Predictor~(MCP) to predict the final inflow count based on the representations.
The locator is trained independently with the coordinate annotations, and existing image crowd localization networks such as FIDT~\cite{FIDT} can be used.
The encoder is trained with our Weakly Supervised Representation Learning method~(WSRL) to extract the discriminative features of each individual. It can be any feature extractor, such as ConvNext~\cite{liu2022convnet}.
MCP does not need training.

\subsection{Weakly Supervised Representation Learning}
\label{sec:sml}
Provided two $\delta$-adjacent frames  $\mf{V}^i$ and $\mf{V}^{j}$ and the corresponding annotations $\mf{P}^i$, $\mf{P}^j$, $\mf{O}^i$ and $\mf{I}^j$, we group the individuals into four sets: $\mf{X}=\{p_u^i|\mf{O}^i_u=0\}$,  $\mf{Y}=\{p_u^j|\mf{I}^j_u=0\}$, denoting shared individuals in the previous frame and the current frame, respectively; $\hat{\mf{X}}=\mf{P}^i-\mf{X}$, and $\hat{\mf{Y}}=\mf{P}^j-\mf{Y}$, denoting outflow individuals and inflow individuals, respectively.
An encoder learns representations $\mf{F}$ for each pedestrian based on the crops around those coordinates. Specifically, to generate the $u$-th feature $f^i_u$ in $\mf{F}$, if the locator outputs points, we extract a rectangle patch of size $96\times64$ centered at the predicted coordinate $p^i_u$, and if the outputs are bounding boxes, we directly crop the bounding box from the original picture. Each cropped patch is resized to $224\times 224$ and fed into the encoder, a ConvNext-S~\cite{liu2022convnet}, generating a feature of $7\times7\times 768$, which is flattened and normalized into a 1D vector $f^i_u$ to represent the final feature for the individual:
\begin{equation}
\begin{aligned}
\mf{F}^i_{\mf{X}}&=\text{Encoder}(\mf{X}), \mf{F}^i_{\hat{\mf{X}}}&=\text{Encoder}(\hat{\mf{X}}) \\
\mf{F}^j_{\mf{Y}}&=\text{Encoder}(\mf{Y}), \mf{F}^j_{\hat{\mf{Y}}}&=\text{Encoder}(\hat{\mf{Y}})
\end{aligned}
\end{equation}
Then, a similarity matrix $\mf{S}\in \mb{R}^{n_i\times n_j}$ between $\mf{F}^i=\left [ \mf{F}^i_{\mf{X}}, \mf{F}^i_{\hat{\mf{X}}} \right ]  $ and $\mf{F}^j=\left [ \mf{F}^j_{\mf{Y}}, \mf{F}^j_{\hat{\mf{Y}}} \right ] $ is computed by:

%
\begin{equation}
\begin{aligned}
     \mf{S}&= 
             \begin{bmatrix}
        (\mf{F}^i_{\mf{X}})^\top \mf{F}^j_{\mf{Y}} & (\mf{F}^i_{\mf{X}})^\top\mf{F}^j_{\hat{\mf{Y}}}  \\
        (\mf{F}^i_{\hat{\mf{X}}})^\top \mf{F}^j_{\mf{Y}}   & (\mf{F}^i_{\hat{\mf{X}}})^\top \mf{F}^j_{\hat{\mf{Y}}}  \\
        \end{bmatrix}=
        \begin{bmatrix}
        \mf{S}_0 & \mf{S}_1  \\
        \mf{S}_2   & \mf{S}_3  \\
        \end{bmatrix}.
\end{aligned}
        \label{eq:s}
        \end{equation}
The matrix $\mf{S}$ is divided into four parts: $\mf{S}_0\in \mb{R}^{m_{i,j}\times m_{i,j}}$, $\mf{S}_1\in \mb{R}^{m_{i,j}\times (n_j-m_{i,j})}$, $ \mf{S}_2\in \mb{R}^{(n_i-m_{i,j})\times m_{i,j}}$, $ \mf{S}_3\in \mb{R}^{(n_i-m_{i,j})\times (n_j-m_{i,j})}$, where $m_{i,j}=(\mf{1}-\mf{I^j})^T(\mf{1}-\mf{O}^i)$ denotes the number of shared individuals in $\mf{V}^i$ and $\mf{V}^j$. 
$\mf{S}_0$ is the similarity matrix of $\mf{X}$ and $\mf{Y}$, while $\mf{S}_1$, $\mf{S}_2$ and $\mf{S}_3$ are the similarity matrix of $\mf{X}$ and $\hat{\mf{Y}}$, of $\hat{\mf{X}}$ and $\mf{Y}$, of $\hat{\mf{X}}$ and $\hat{\mf{Y}}$, respectively.
%
%

To pull the matched groups ($\mf{X}$ and $\mf{Y}$) closer and push away individual pairs from unmatched groups, 
we propose a weakly supervised Group-level Matching Loss~($\textbf{GML}$) to constrain different parts of $\mf{S}$.

\vspace{1mm}
\noindent\textbf{Constraint for $\mf{S}_0,\mf{S}_1,\mf{S}_2$.} $\mf{S}_0$ denotes the similarity among individuals shared between two frames, and it does not matter whether the individuals are correctly matched as long as we can assign them a one-to-one match. As for $\mf{S}_1$ and $\mf{S}_2$, they are the similarities of different identities and thus should be zero.
 
To this end, we introduce a latent variable matrix $\Omega \in \mb{R}^{m_{i,j}\times m_{i,j}}$, where the $(u,v)$-th element $\Omega_{u,v}$ denote the probability that the $u$-th pedestrian in $\mf{X}$ is matched with $v$-th pedestrian in $\mf{Y}$. To increase the similarity between the pairs that have a higher matching probability and decrease the similarity between the pairs that have a lower matching probability, inspired by \cite{infonce,cwcl}, we define the soft contrastive loss as:
\begin{equation}
\begin{aligned}
\mc{L}_{scon}(i)&=\mathop{\min}_{\quad\Omega} -\sum_{u=1}^{m_{i,j}}\sum_{v=1}^{m_{i,j}}\Omega_{u,v}C_{u, v},\\
s.t. \text{\quad} & \mf{1}_{m_{i,j}}^{T}{\Omega}=\mf{1}, \text{\quad}
        {\Omega}\mf{1}_{m_{i,j}}=\mf{1}, 0\leq\Omega_{u,v}\leq1,
\end{aligned}
\label{eq:f} 
\end{equation}
where $C_{u, v}$ 
denotes the contrastive similarity ~\cite{chen2020mocov2},
and is calculated as,
\begin{equation}
\resizebox{0.43\textwidth}{!}{$
    C_{u, v}=\frac{e^{\frac{1}{\gamma}  \mf{S}_{u,v}}}{e^{\frac{1}{\gamma} \mf{S}_{u,v}}+\sum\limits^{u'\neq u}_{1\leq u'\leq n_i}e^{\frac{1}{\gamma} \mf{S}_{u',v}}+\sum\limits^{v'\neq v}_{1\leq v'\leq n_j}e^{\frac{1}{\gamma}  \mf{S}_{u,v'}}},$}
    \label{eq:cost}    
\end{equation}
where $\gamma$ is a temperature hyper-parameter~\cite{nonparameter} and $\mf{S}_{u,v}=\langle f^i_u,f^j_v \rangle$ is the $(u,v)$-th element of $\mf{S}$, calculated by the $u$-th feature in $\mf{F}^i$ and the $v$-th feature in $\mf{F}^j$.
As $\Omega_{u,v}$ describes the matching probability, our soft contrastive loss evaluates the expectation of contrastive loss.
It should be noting that in $C_{u, v}$, $(f_u^i, f_v^j)$ is considered as positive pairs, while $\{(f_u^i, f_{v'}^j)|v'\neq v\}$ and $\{(f_{u'}^i, f_v^j)| u'\neq u\}$ are considered as negative pairs.
Thus, $\mc{L}_{scon}$ can encourage $\mf{S}_0$ to be a permutation matrix, which indicates that there exists a one-to-one match between $\mf{X}$ and $\hat{\mf{Y}}$, and drive $\mf{S}_1,\mf{S}_2$ to be zero matrices.

The problem defined by \cref{eq:f} is a typical balanced Optimal Transport~(OT~\cite{ot}) problem which transports $\mf{1}_{m_{i,j}}$ to $\mf{1}_{m_{i,j}}$ using $\mf{1}-C$ as the cost matrix and $\Omega$ is the transport matrix. Therefore, we consider $\mc{L}_{scon}(i)$ as an OT loss, which can be solved with the Sinkhorn algorithm~\cite{sinkhorn}. 

\noindent\textbf{Constraint for $\mf{S}_3$. } Obviously, the elements in $\mf{S}_3$ measure the similarity of different individuals, so they should be as small as possible. However, directly constraining $\mf{S}_3$ to zero matrix will make features from $\hat{\mf{X}},\hat{\mf{Y}}$ easily collapse to zero-vector. To this end, the widely used Hinge L1 loss \cite{hinge} is adopted to constrain $\mf{S}_3$:
\begin{equation}
    \mathcal{L}_{hinge}(i)=\frac{\sum Relu(\mf{S_3}-\theta)}{(n_i-m_{i,j})(n_j - m_{i,j})},
    \label{eq:hinge}
\end{equation}
where $\theta$ is a threshold used to ignore small values in $\mf{S}_3$.

Finally, the group-level matching loss of the training set $\mf{V}$ is formulated as,
\begin{equation}
    \mf{GML}(\mf{V})=\sum\limits_{0<i<T,i=k\delta+1}\left(\mathcal{L}_{scon}(i)+\mathcal{L}_{hinge}(i)\right).
\end{equation}

\subsection{Memory-based Count Predictor}
\label{sec:predictor}
To handle individuals' appearance variance and re-entering cases, inspired by~\cite{template1,template2,template3}, we propose a memory-based count predictor~(MCP) to reason the final count for the video, which stores templates of recently appeared individuals in a flexible memory. Specifically, when processing frame $\mf{V}^{i+\delta}$, the memory is denoted as $\mc{G}^i=\{\mc{G}^i_1,\mc{G}^i_2,..., \mc{G}^i_{\mf{N}^i}\}$, where $\mf{N}^i$ is the memory size at time $i$. $\mc{G}^i_u=\{ g^i_{u,0}, ..., g^i_{u,k} \}$ is a set of templates for $u$-th pedestrian in the memory, and $k$ is the number of templates stored for $u$-th pedestrian.

We first use the crowd locator to predict the coordinates of pedestrians in $\mf{V}^{i+\delta}$, and then extract their corresponding features, denoted as $\{f_1^{i+\delta}, f_2^{i+\delta},\cdots,f_{n_{i+\delta}}^{i+\delta} \}$. To associate the individuals in $\mf{V}^{i+\delta}$ with templates in memory, 
we first define the cost $\hat{C}_{u,u'}$ of matching the $u$-th pedestrian with the $u'$-th template in $\mc{G}^{i}$ as
\begin{equation}
    \hat{C}_{u,u'}=\mathop{\max}_{v}(1- \langle f^{i+\delta}_{u}, g^i_{u',v} \rangle ).
\end{equation}
Then, we generate an optimal one-to-one match $\pi \in \{0,1\}^{n_{i+\delta} \times \mf{N}^i}$~(assuming $\mf{N}^i\geq n_i$) by solving the following problem with Hungarian Algorithm~\cite{lap}: 
\begin{equation}
\begin{aligned}
    \pi &= \mathop{\arg \min}_{\pi} \sum\limits_{u,u'}\pi_{u,u'}\hat{C}_{u,u'}\\
    s.t. \text{\quad} & \forall u', \sum\limits_{u}{\pi}_{u,u'}=1, \text{\quad}
        \forall u, \sum\limits_{u}{\pi}_{u,u'}\leq1.
\end{aligned}
\label{eq:hungry}
\end{equation}
If the matching cost for $u$ meets $\sum_{u'=1}^{\mf{N}^i}\pi_{u,u'}\hat{C}_{u,u'}> \zeta$, $u$ will be judged as an inflow pedestrian. Otherwise, if $\pi_{u,u'}=1$, $u$ is associated with $g^i_{u'}$.

\noindent{\textbf{Template update.}} Similar to \cite{template1}, $\mc{G}^i_u$ has a time-to-live factor~($\textbf{ttl}$). $\textbf{ttl}_{u}$ will decrease one if $\mc{G}^i_u$ has no mapping to any pedestrian feature in $\mf{F}^{i+\delta}$. If $\textbf{ttl}_{u}$ decreases to zero, the template $\mc{G}^i_u$ will be dropped. Meanwhile, whenever $\mc{G}^i_u$ is associated with a pedestrian in the current frame, $\textbf{ttl}_{u}$ will be reset to $\textbf{ttlmax}$. This time-to-live threshold allows us to save some recently appeared targets while discarding those that have not appeared for a long time. If pedestrian $u$ is associated with $u'$-th template, we add $f^{i+\delta}_{u}$ to $\mc{G}^i_{u'}$ to update the template
\begin{equation}
\resizebox{0.42\textwidth}{!}{$\mc{G}^{i+\delta}_{u'}\gets \mc{G}^{i}_{u'}\cup\{f^{i+\delta}_{u}|\pi_{u,u'}=1, \sum_{u'=1}^{\mf{N}^i}\pi_{u,u'}\hat{C}_{u,u'}\leq \zeta \}.$}
\end{equation}
If pedestrian $u$ is an inflow pedestrian, $\{f^{i+\delta}_{u}\}$ will be added to the memory $\mc{G}^{i+\delta}$ as a new template.

The output number of the total individuals in video $I$ is then calculated via
\begin{equation}
\hat{N}=n_1+\mathop{\sum}_{\delta+1\leq i \leq T, i=k\delta+1}\hat{n}_i,
\end{equation}
where $\hat{n}_i$ is the number of inflow pedestrians at time $i$.

\section{Experiments}
\label{sec:exp}

\subsection{Dataset}
We test our CGNet on three datasets for experimental comparison: CroHD~\cite{crohd}, SenseCrowd~\cite{sensecrowd}, and UAVVIC.
CroHD has four training videos and five testing videos. 
SenseCrowd contains 634 videos, and we split the train, validation, and test dataset following DRNet~\cite{DRVIC}. UAVVIC is our proposed dataset collected by a moving UAV camera in various scenes, including campus, beach, car park, highway, city road, and square. UAVVIC consists of 221 videos (100 for training, 100 for testing, and 21 for validation), and 5,396 frames are sampled with 3s as the interval. Annotations consist of 398,158 bounding boxes, and group-level matching labels in neighbor frames are provided. The resolutions of UAVVIC are 4K and 1080P for better capture of tiny pedestrians from drone view. A detailed comparison of UAVVIC with existing video crowd counting datasets~\cite{mall,ucsd,fdst,sensecrowd,crohd} is shown in \cref{tab:dataset}. UAVVIC provided a moving shot with a larger range of pedestrians. More details about UAVVIC are provided in the supplementary materials.

\subsection{Metrics}
Mean Absolute Error~(MAE), Mean Square Error~(MSE), and Weighted Relative Absolute Errors~(WRAE) are used for evaluation. The first two metrics are common metrics applied in VCC. 
However, unlike VCC schemes, we only count the same person once in the video. 
WRAE~\cite{DRVIC} is proposed to balance the performance on videos with different lengths and pedestrian numbers:
\begin{equation}
    WRAE=\sum_{i=1}^{K}\frac{T_i}{\sum_{j=1}^{K}T_j}\frac{|N_i-\hat{N}_{i}|}{N_i},
\end{equation}
where $K$ is the total number of videos, and $T_i$ is the length of video $i$, $N_i/\hat{N}_i$ denotes the number of ground truth/predicted counts in $i$-th video. 
\begin{table}[!bt]
    \centering
    \small
    \resizebox{\linewidth}{!}{ 
    \begin{tabular}{c c cc ccc cc}
    \hline
         Name &Resolution &Range& Moving shot & Point & Box &Trajectory & In-Out  \\\hline
         Mall~\cite{mall} & 640*480  & 13-53& \XSolidBrush & \Checkmark & \XSolidBrush & \XSolidBrush & \XSolidBrush \\
         UCSD~\cite{ucsd}& 238*158 &11-46& \XSolidBrush  & \Checkmark & \XSolidBrush &\XSolidBrush & \XSolidBrush\\
         FDST~\cite{fdst} & \makecell[c]{1920*1080 \\ 1280*720} & 9-57  & \XSolidBrush & \Checkmark & \XSolidBrush &\XSolidBrush & \XSolidBrush\\ \hline
         SenseCrowd~\cite{sensecrowd}& - &1-296& \XSolidBrush  & \Checkmark & \Checkmark &\Checkmark & \XSolidBrush\\
         CroHD~\cite{crohd}  & \makecell[c]{-} &25-346& \XSolidBrush& \Checkmark & \Checkmark &\Checkmark  & \XSolidBrush\\
         UAVVIC & \makecell[c]{3840*2160 \\ 1920*1080} & 0-735 & \Checkmark  & \Checkmark  & \Checkmark &  \XSolidBrush & \Checkmark\\
    \hline
    \end{tabular}
    }
    \caption{Comparison of different video crowd counting datasets.}
    \label{tab:dataset}
    \vspace{-5mm}
\end{table}

\begin{figure*}[!hbt]
\begin{subfigure}{\textwidth}
\centering
\includegraphics[width=0.92\linewidth]{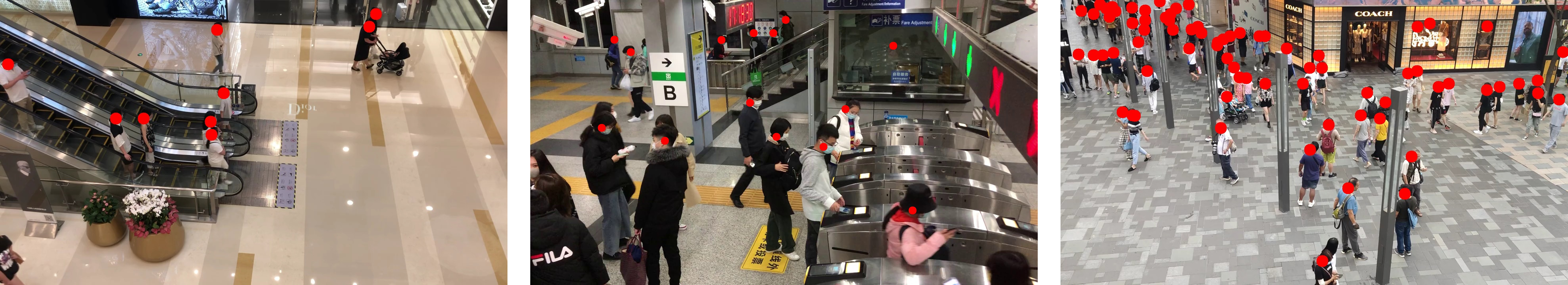}
\caption{Previous frame~(0s)}
\label{fig:q1}
\end{subfigure}

\begin{subfigure}{\textwidth}
\centering
\includegraphics[width=0.92\linewidth]{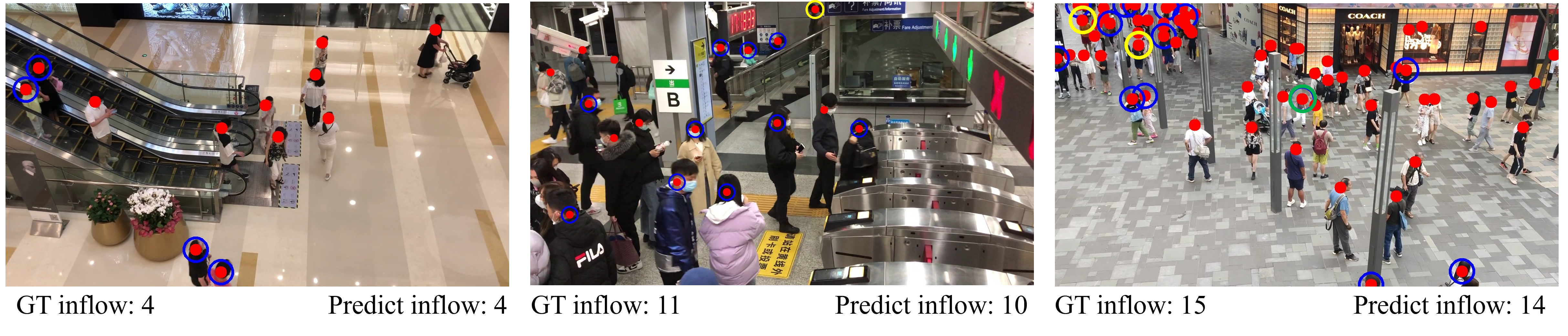}
\caption{Current frame~(3s)}
\label{fig:q2}
\end{subfigure}
\vspace{-3mm}
\caption{Results on the SenseCrowd dataset. Red dots are predictions of the locator. Blue/green/yellow circles are correct predicted inflow, error predicted inflow, and missed inflow, respectively. }
\label{fig:quality}
\vspace{-3mm}
\end{figure*}

\begin{table*}[!hbt]
    \centering
    \small
    { 
    \begin{tabular}{c ccc ccccc}
    \hline
        \multirow{2}{*}{Method} &\multirow{2}{*}{~~MAE~~} & \multirow{2}{*}{~~~MSE~~} & \multirow{2}{*}{~~WRAE(\%)~~} & \multicolumn{5}{c}{MAE on five different density levels}\\\cmidrule(r){5-9}
        &&&&~~~~~D0~~~~ &~~~~~D1~~~~ &~~~~~D2~~~~ & ~~~~~D3~~~~ & ~~~~~D4~~~~   \\\hline
        FairMOT~\cite{fairmot} & 35.4 & 62.3 & 48.9 & 13.5 & 22.4 & 67.9 & 84.4 & 145.8 \\
        HeadHunter-T~\cite{headhunter} & 30.0 & 50.6 & 38.6 &11.8 & 25.7& 56.0 & 92.6 & 131.4\\
        SMILE~\cite{smile} &27.22 & 36.7 & 32.5 & 9.2 & 21.0 & 33.8 & 76.5 & 203.5\\
        SparseTrack~\cite{sparsetrack} & 28.36 & 43.8 & 30.8 & 9.9 & 21.5 & 42.1 & 89.2 & 189.5\\
        Deep-OC-SORT~\cite{deepocsort} & 26.04 & 43.6 & 29.5 & 9.5 & 25.1 & 29.1 & 60.3 & 156.5 \\
        BoT-SORT~\cite{botsort} & 24.67 & 32.4 & 28.3 & 8.7 & 20.5 & 31.1 & 70.2 & 165.1 \\\hline
        LOI~\cite{zhao2016crossing} & 24.7 & 33.1 & 37.4 & 12.5 &  25.4 & 39.3 & 39.6 & 86.7\\
        DRNet~\cite{DRVIC} & 12.3 & 24.7 & 12.7 & \textcolor{blue}{4.1} & 8.0 & 23.3 &  50.0 & 77.0\\\hline
        CGNet~(Ours) 
         & \textcolor{blue}{8.86} & \textcolor{blue}{17.69} & \textcolor{blue}{12.6} & 5.0 & \textcolor{blue}{5.8} & \textcolor{blue}{8.5} & \textcolor{blue}{25.0} & \textcolor{blue}{63.4}
         \\\hline
    \end{tabular}
    }
    \caption{Performance comparison on SenseCrowd. The density labels D0-D4 are defined in \cite{DRVIC}, where D0, D1, D2, D3, and D4 correspond to the count range [0, 50), [50, 100), [100, 150), [150, 200) and [200, +$\infty$), respectively.
    The best values are highlighted in blue font. }
    \label{tab:sensecrowd}
     \vspace{-2mm}
\end{table*}

\begin{table*}[!bt]
    \centering
    \small
    { 
    \begin{tabular}{c ccc ccccc}
    \hline
                \multirow{2}{*}{Method} &\multirow{2}{*}{MAE} & \multirow{2}{*}{MSE} & \multirow{2}{*}{WRAE(\%)}  & \multicolumn{5}{c}{MAE on five testing scenes}\\\cmidrule(r){5-9}
        &&& &CroHD11 & CroHD12 & CroHD13 & CroHD14 & CroHD15   \\\hline
        PHDTT~\cite{phdtt} & 2130.4 & 2808.3 & 401.6 & 247 & 3793 & 4794 & 491 & 1327\\
        FairMOT~\cite{fairmot} & 256.2 & 300.8 & 44.1 &11&427&284&408&1000\\
        HeadHunter-T~\cite{headhunter} & 253.2 & 351.7 & 32.7  &65&101&515&582&1003 \\
        SMILE~\cite{smile} &257.6 & 334.5 & 40.5 &16&231&156&649&246\\
        SparseTrack~\cite{sparsetrack} & 176.6 & 208.6 & 27.6 &29&172&258&336&88\\
        Deep-OC-SORT~\cite{deepocsort} & 165.2 & 195.9 & 33.1 &68&351&186&161&60 \\
        BoT-SORT~\cite{botsort} & 154.8 & 176.42 & 27.4 &49&210&138&286&91 \\\hline
        LOI~\cite{zhao2016crossing} & 305.0 & 371.1 & 46.0 & 60&243&458&630&131\\
        DRNet~\cite{DRVIC} & 141.1 & 192.3 & 27.4 &31&338&18&255 & \textcolor{blue}{61}\\\hline
        CGNet~(Ours) 
         & \textcolor{blue}{75.0} & \textcolor{blue}{95.1} & \textcolor{blue}{14.5}&\textcolor{blue}{7}&\textcolor{blue}{72}&\textcolor{blue}{14}&\textcolor{blue}{144}&138
         \\\hline
    \end{tabular}
    }
    \caption{Performance comparison on CroHD dataset. CroHD11-CroHD15 are five test scenes labeled in \cite{DRVIC}. The best values are highlighted in blue font. }
    \label{tab:CroHD}
\vspace{-3mm}
\end{table*}

\begin{table*}[!bt]
    \centering
    \small
    \begin{tabular}{c c|cccc }
    \hline
         \multirow{2}{*}{Method} &\multirow{2}{*}{MAE/MSE/WRAE} &\multicolumn{4}{c}{MAE/MSE/WRAE on four different categories}\\\cmidrule(r){3-6}
         &  & Pedestrian & Car&Bus&Van\\\hline
         BoT-SORT~\cite{botsort} & 49.9/228.2/29.9&92.4/398.2/49.0&59.1/259.1/39.2&23.0/109.6/57.1&25.1/146.2/77.3\\
         Deep-OC-SORT~\cite{deepocsort} & 37.1/115.4/35.2&85.2/213.5/22.1&33.4/113.1/21.0&19.1/~~78.2/26.2&10.7/56.6/32.8\\
         DRNet~\cite{DRVIC} & 18.4/~41.0/15.2&34.5/\textcolor{blue}{~~92.1}/19.1&19.0/~~32.7/17.8&10.4/~~22.1/32.2&~9.9/17.2/29.4\\\hline
         CGNet~(Ours)& \textcolor{blue}{12.9}/\textcolor{blue}{~~37.4}/\textcolor{blue}{~~12.0}&\textcolor{blue}{31.2}/~~98.4/\textcolor{blue}{14.4}&\textcolor{blue}{16.9}/\textcolor{blue}{~~45.2}/\textcolor{blue}{~7.3}&\textcolor{blue}{~2.1}/\textcolor{blue}{~~~~4.1}/\textcolor{blue}{19.2}&\textcolor{blue}{~~1.3}/\textcolor{blue}{~~2.0}/\textcolor{blue}{25.1} \\
    \hline
    \end{tabular}
     \caption{Performance comparison on UAVVIC. All the methods are trained independently with four classes. The values of each entry are MAE/MSE/WRAE. The best values are highlighted in blue font. }
     \label{tab:uavvic}
\end{table*}
\subsection{Implementation Details}
\label{sec:detail}
\noindent\textbf{Training.} If there are no special instructions, the locator we applied is a well-known crowd localization model, FIDT~\cite{FIDT}. All parameters are official from FIDT, except that we replaced the backbone from HRNet-W48~\cite{hrnet} to HRNet-W18~\cite{hrnet} for efficient inference.
%
%
%
The learning rate is set as $1e^{-4}$ along with AdamW~\cite{adamw} as the optimizer and applying the pre-train weight from Timm~\cite{rw2019timm}. All related models can be trained on one RTX3090~(24G memory).

\noindent\textbf{Testing.} In the testing phase, the time interval $\delta$ is set as $3s$ following the setting in \cite{DRVIC}, the threshold $\zeta$ is set as $0.7$, $\lambda$ in \cref{eq:cost} is set as $10$, and the max time-to-live~($\textbf{ttlmax}$) is set as 3,  the max size of memory~($\textbf{memmax}$) is set as $5$. 
\subsection{Overall Performance}

\begin{table*}[!hbt]
    \centering
    \small
    { 
    \begin{tabular}{cc ccc| ccccc}
    \hline
         \multirow{2}{*}{~~\#~~}&\multirow{2}{*}{Method} &\multirow{2}{*}{~~~MAE~~~} & \multirow{2}{*}{~~~MSE~~~} & \multirow{2}{*}{~~~WRAE(\%)~~~~} & \multicolumn{5}{c}{~~~MAE on five different density levels~~~~}\\\cmidrule(r){6-10}
        &&&&&~~~D0~~~&D1 & D2 & D3 & D4   \\\hline
         \ding{172} &No constrain on $\mf{S}_1,\mf{S}_2$& 13.35 & 32.3 & 22.8& 9.7& 12.3&15.3& 26.5 &43.4\\
         \ding{173} &w/o \cref{eq:hinge}& 9.84 & 20.0 & 13.8 & 5.3 & 6.3 & 12.0 & 28.8 & 67.7 \\
         \ding{174} & w/o MCP & 10.71 & \textcolor{blue}{17.40} & 16.9 &6.9 &8.9&11.7 &\textcolor{blue}{24.3} & \textcolor{blue}{50.6} \\
         \ding{175}  & full CGNet & \textcolor{blue}{8.86} & 17.69 & \textcolor{blue}{12.6} & \textcolor{blue}{5.0} & \textcolor{blue}{5.8} & \textcolor{blue}{8.5} & 25.0 & 63.4 \\
    \hline
    \end{tabular}
    }
    \caption{Ablation study on the main components. The density labels D0-D4 are defined in \cite{DRVIC}, where D0, D1, D2, D3, and D4 correspond to the count range [0, 50), [50, 100), [100, 150), [150, 200) and [200, +$\infty$), respectively. The best values are highlighted in blue font.}
    \label{tab:mainaba}
     \vspace{-4mm}
\end{table*}

\noindent\textbf{Comparison Methods.} To the best of our knowledge, only DRNet\cite{DRVIC} has been specifically designed for the VIC task. To evaluate the effectiveness of our proposed CGNet, we also tested two categories of approaches that could be applied to VIC: 1) Multi-object tracking (MOT) methods, including HeadHunter-T~\cite{headhunter}, FairMOT~\cite{fairmot}, PHDTT~\cite{phdtt}, SMILE~\cite{smile}, SparseTrack~\cite{sparsetrack}, Deep-OC-SORT~\cite{deepocsort}, and BoT-SORT~\cite{botsort}, where we consider the total number of indices in the entire video as individual count.
We set the frame rate to 1 FPS for HeadHunter-T and 0.33 FPS for the other MOT methods to obtain better VIC performance. 2) A recent cross-line video crowd-counting, LOI~\cite{zhao2016crossing}, which also requires trajectory labels in training.

\noindent\textbf{Results on SenseCrowd:} As shown in \cref{tab:sensecrowd}, it is evident that our CGNet exhibits superior performance on this larger dataset, SenseCrowd. Particularly, CGNet improved the overall MAE by about 28\% compared to DRNet. Furthermore, we achieve the lowest MAE across all density levels except for D0. 
Compared to the existing lowest MAE on D1, D2, D3, and D4, the performance improvements of our CGNet are about 27\%, 63\%, 37\%, and 17\%, respectively.
In \cref{fig:quality}, we show qualitative results on three representative scenes,i.e., mall, station, and outdoor plaza. Although the inflows in the current flow are not concentrated, CGNet can accurately predict them.
More visualization results can be found in supplementary material.

\noindent\textbf{Results on CroHD.} Comparison with existing methods on CroHD is shown in \cref{tab:CroHD}. Our CGNet achieved the lowest MAE, MSE, and WRAE. Especially, except for CroHD15, our method obtained the lowest MAE across the other four testing scenes. 
This may be because the resolution of CroHD15 is $1920\times734$, while that of training videos is $1920\times1080$. As CGNet is trained with only point annotations, it cannot crop the patch of optimal size to extract features on CroHD15.
It should be noted that on CroHD13 and CroHD14, which have the highest average density~(245.9 and 259.6 person/frame), the performances of all the compared methods except for DRNet degrade significantly. In contrast, the performance of our CGNet is relatively stable in these two scenes. 
Furthermore, even without individual-level matching labels, our CGNet performs favorably against the state-of-the-art approach, DRNet, by a large margin (about 46\% on the MAE),  demonstrating the effectiveness of our method.

\noindent\textbf{Results on UAVVIC.}
The results on our proposed UAVVIC dataset are shown in \cref{tab:uavvic}. All methods are trained and tested on each category, respectively. 
Since there is no trajectory annotation on UAVVIC, we train the compared methods \cite{botsort,deepocsort,DRVIC} on VisDrone~\cite{visdrone}, a UAV dataset for MOT. 
Our CGNet shows favorable performance in all categories. The average MAE is improved by about 30\% compared to DRNet.

\subsection{Ablation Studies}

We conducted several ablation studies on the SenseCrowd.
1). We verify the effectiveness of the main components of our CGNet.
2). We analyze the effect of two main parameters.
3). We report the performance of our CGNet with different locators.
4). We use the trajectories generated by our CGNet as pseudo-trajectory labels for existing VIC methods.
5). We compare the time cost of the WVIC and VIC labels.

\noindent\textbf{Effectiveness of Constrain for $\mf{S}_1, \mf{S}_2$.} \cref{eq:cost} constrains $\mf{S}_0,\mf{S}_1,\mf{S}_2$ together. Without considering un-matched pairs in $\mf{X} \times \hat{\mf{Y}}$ and $\hat{\mf{X}} \times\mf{Y}$, it can be changed to constrain only $\mf{S}_0$, where \cref{eq:cost} becomes:
\begin{equation}
\resizebox{0.42\textwidth}{!}{ $C_{u, v}=\frac{e^{\gamma \mf{S}_{u,v}}}{e^{\gamma \mf{S}_{u,v}}+\sum\limits^{u'\neq u}_{1\leq u'\leq n_i}e^{\gamma \mf{S}_{u',v}}+\sum\limits^{v'\neq v}_{1\leq v' \leq n_j}e^{\gamma \mf{S}_{u,v'}}}.$
}
\end{equation}
As shown in \cref{tab:mainaba}, MAE dropped by about 4.51 without constraints on $\mf{S}_1$ and $\mf{S}_2$ (\ding{172} VS \ding{175}). 
This is because elements in $\mf{S}_1$ and $\mf{S}_2$ are related to pedestrians that exist only in one frame.
Considering $\mf{X} \times \hat{\mf{Y}}$ and $\hat{\mf{X}} \times\mf{Y}$ as negative pairs, our soft contrastive loss can push apart individuals in $\hat{\mf{X}}$ ($\hat{\mf{Y}}$) with that in $\mf{Y}$ ($\mf{X}$), thus enhancing the distinguishing ability of the learned representations.

\noindent\textbf{Effectiveness of \cref{eq:hinge}.} We replace the hinge MAE loss in \cref{eq:hinge} to normal L1 loss and derive the method \ding{173} in \cref{tab:mainaba}.
Compared to L1 loss, our $\mathcal{L}_{hinge}(\cdot)$ gives a soft margin to the inner product of two features other than encouraging them to be 0. 
Optimizing the inner product of all features toward 0 makes it easier to cause either side to collapse a $\mf{0}$ vector, which is not conducive to learning different representations for each individual. 
Compared the performance of \ding{173} with that of \ding{175} in \cref{tab:mainaba}, $\mathcal{L}_{hinge}(\cdot)$ improves MAE in all density levels. 
\begin{figure}[!bt]
     \centering
     \begin{subfigure}[b]{0.22\textwidth}
         \centering
         \includegraphics[width=\textwidth,height=4.5cm]{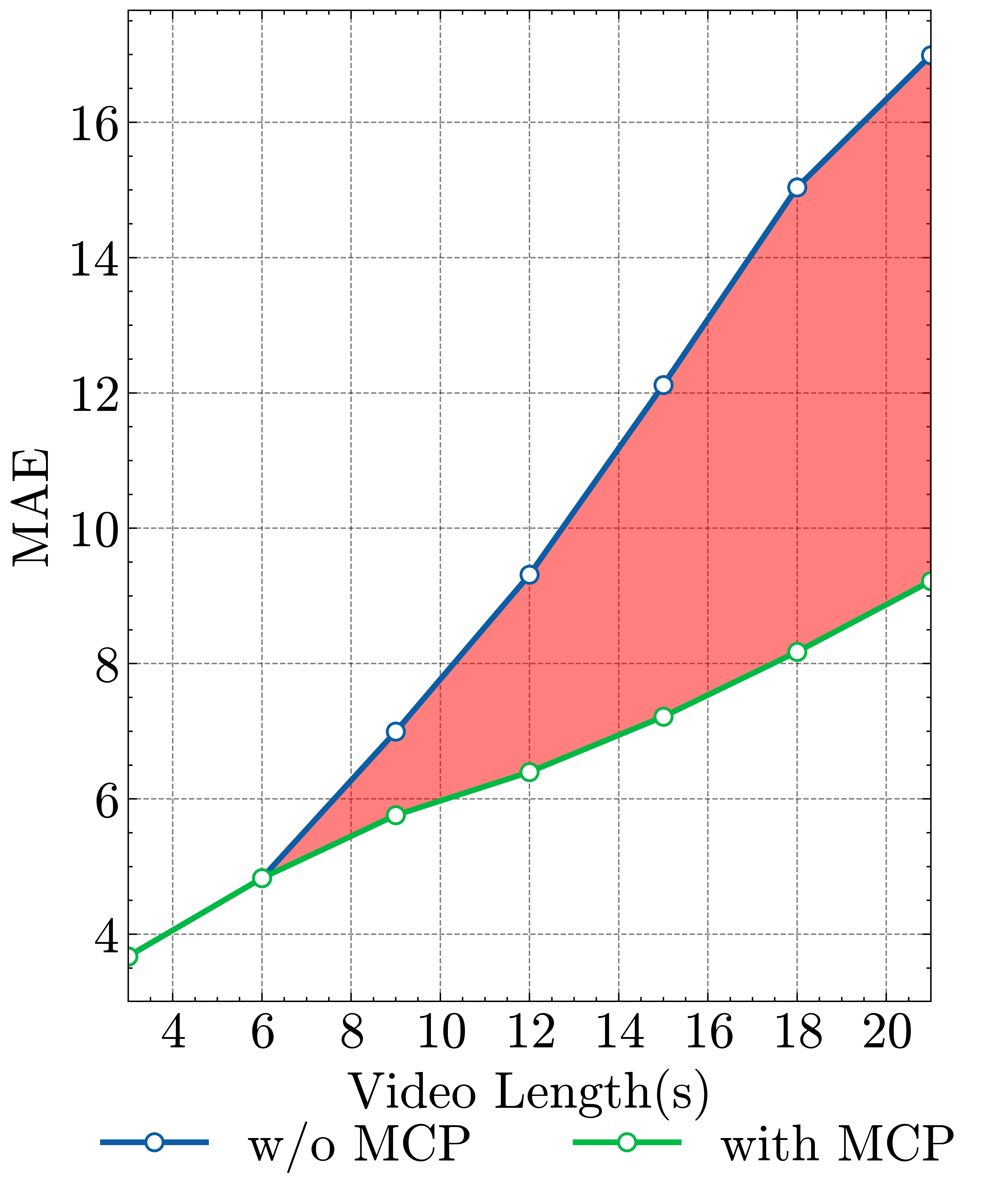}
         \caption{}
         \label{fig:3a}
     \end{subfigure}
     \begin{subfigure}[b]{0.22\textwidth}
         \centering
         \includegraphics[width=\textwidth,]{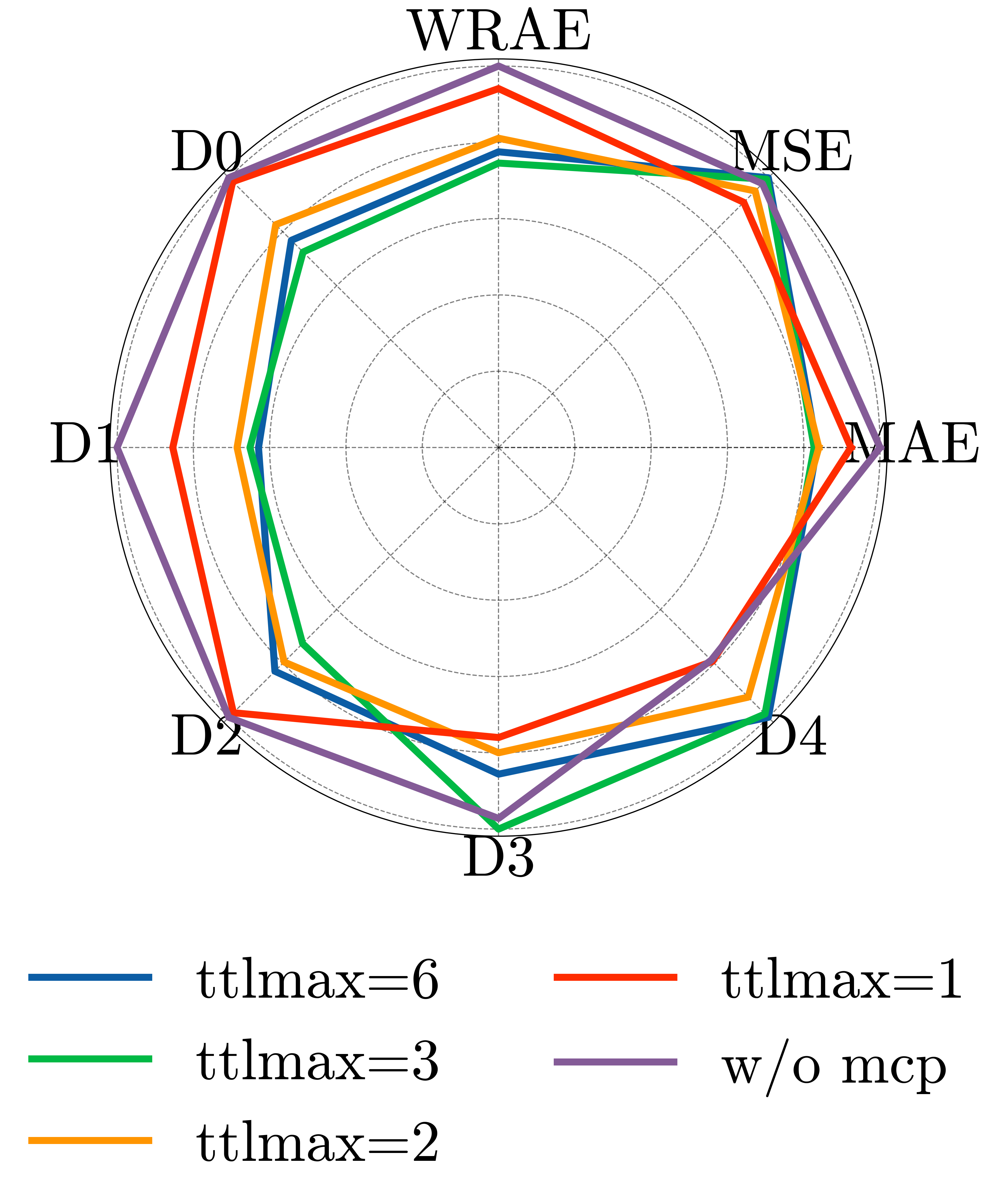}
         \caption{}
         \label{fig:3b}
     \end{subfigure}
    \caption{Effectiveness of MCP on SenseCrowd. (a) The average MAE with different lengths of videos with and w/o MCP. (b) Performance with different max time-to-live factor \textbf{ttlmax} of MCP.}
    \label{fig:mcp}
     \vspace{-4mm}
\end{figure}

\noindent\textbf{Effectiveness of MCP.}
As shown in \cref{tab:mainaba}, compared \ding{174} with \ding{175}, MCP brings about 4.1 improvements on WRAE.
Specifically, we evaluate the average MAE of our CGNet on videos with different lengths.
As shown in \cref{fig:mcp}(a), the performance gap between \ding{174} and \ding{175} becomes larger as the length of the video increases.
This is because, with the stored templates of individuals, MCP can handle appearance variations and individuals' re-entry, alleviating error accumulation during inflow reasoning.

\noindent\textbf{Effect of $\textbf{ttlmax}$.} 
The performance with different max long-time-live factor $\textbf{ttlmax}$ is shown in \cref{fig:mcp}(b). 
%
Compared with no memory (i.e., $\textbf{ttlmax}=0$), adding memory is helpful for all density levels except for extremely high-density levels (D4).
It may be because high-density crowds result in too many candidate individual-level matches, making it difficult for our CGNet to choose the right one. 
Meanwhile, the performance degrades when $\textbf{ttlmax}$ increases from 3 to 6. 
This is because larger $\textbf{ttlmax}=6$ leads to the templates of individuals that appeared a long time ago being stored in the memory.
As the probability of these people re-entering the field is very low, the recorded templates become distractions.

\begin{table}[!bt]
    \centering
    \small
    \resizebox{\linewidth}{!}{ 
    \begin{tabular}{cc ccc ccccc}
    \hline
            \multirow{2}{*}{$\delta$} &\multirow{2}{*}{MAE} & \multirow{2}{*}{MSE} & \multirow{2}{*}{WRAE(\%)} & \multicolumn{5}{c}{MAE on five different density levels}\\\cmidrule(r){5-9}
        &&&&D0 &D1 & D2 & D3 & D4   \\\hline
         5  & 10.96 & 20.26 & 19.5 & 6.5 & 6.9  & 10.4  & 29.2 & 79.9 \\
         4  & \textcolor{blue}{8.71}  & \textcolor{blue}{14.33} & 12.9 & \textcolor{blue}{4.2} & \textcolor{blue}{4.9}  & 10.2  & 27.8 & 70.1  \\
         3  & 8.86  & 17.69 & \textcolor{blue}{12.6} & 5.0 & 5.8  & \textcolor{blue}{8.5}   & 25.0 & 63.4 \\
         2  & 9.77  & 20.01 & 17.3 & 7.1 & 8.8  & 12.4  & \textcolor{blue}{17.4} & 52.6 \\
         1  & 9.56  & 15.34 & 16.6 & 6.8 & 7.4  & 10.5  & 23.1 & \textcolor{blue}{45.5} \\
         
    \hline
    \end{tabular}
    }
    \caption{Ablation study on interval $\delta$. The best values are highlighted in blue font.}
    \label{tab:delta}
    \vspace{-3mm}
\end{table}

\noindent\textbf{Effect of Interval $\delta$.}
The performance with different intervals $\delta$ is shown in \cref{tab:delta}.  The performance on higher-density video is better when $\delta$ is smaller. This is because crowds flow faster in those videos, and long intervals may lead to missing pedestrians in sampled frames.
The overall best performance is achieved when $\delta$ is set as $4s$.
\begin{table}[!bt]
    \centering
    
\resizebox{\linewidth}{!}{ 
    \begin{tabular}{ccc ccc}
    \hline
         Method & Locator & Encoder &MAE & MSE & WRAE  \\\hline
         \ding{172} & FIDT & ConvNext-S&8.86 & 17.69 & 9.27\\
         \ding{173} & Yolov8 nano&ConvNext-S&8.01 & 16.42 & 8.85\\
         \ding{174} & Yolov8 small&ConvNext-S&\textcolor{blue}{7.56} & \textcolor{blue}{15.19} & \textcolor{blue}{8.05}\\
         \ding{175} & VGG16+FPN&ConvNext-S&9.20 & 17.45 & 9.55 \\
         \ding{176} & VGG16+FPN &PrRoIPooling&10.01 & 18.22 & 9.70 \\
         \ding{177} & VGG16+FPN &PrRoIPooling&12.59 & 23.32 & 12.58\\
         \hline
    \end{tabular}
    }
    \caption{Performance on SenseCrowd of our scheme with different locators and feature extractors.
    In \ding{177}, MCP is replaced with the inflow reasoning method in DRNet. The best values are highlighted in blue font.
    }
    \label{tab:locater}
    \vspace{-5mm}
\end{table}

\noindent\textbf{Effect of the Locator.}
We replaced FIDT in our CGNet with Yolo V8~\cite{yolov8_ultralytics} nano and small and the localization branch~(VGG16+FPN) used in DRNet, respectively. 
Different from FIDT, these three locators are trained with box annotations. 
For Yolo V8, the predicted bounding boxes are fed to our feature extractor to generate representations of individuals.
For a fair comparison with DRNet, we also replace our feature extractor ConvNext-S with PrRoIPooling~\cite{proi} used in DRNet~\cite{DRVIC} and remove MCP. 
As shown in \cref{tab:locater}, changing locators can further boost the performance of our method. With the same locator and feature extractor as DRNet~(\ding{176}), the MAE of our method is 10.01, 2.99 lower than that of DRNet.
Even without MCP~(\ding{177}), the MAE of our method is only slightly larger than that of DRNet~(12.59 v.s. 12.3), demonstrating that our method can learn effective representations without trajectory labels.

\begin{table}[!bt]
    \centering
    \small
    \resizebox{\linewidth}{!}{ 
    \begin{tabular}{c ccc| ccc}
    \hline
         \multirow{2}{*}{Locater} & \multicolumn{3}{c|}{Ground Truth Labels} & \multicolumn{3}{c}{Pseudo Labels}\\\cmidrule(r){2-7}
        & MAE & MSE & WRAE & MAE & MSE & WRAE   \\
        \hline
        FairMOT & 35.4 & 62.3 & 48.9 & 37.2 & 64.8 & 49.8 \\
        HeadHunter-T & 30.0 & 50.6 & 38.6 & 32.5 & 54.2 & 39.9 \\
        SMILE &27.2 & 36.7 & 32.5 & 27.5 & 37.0 & 35.2 \\
        SparseTrack & 28.4 & 43.8 & 30.8 & 30.5 & 45.0 & 31.5\\
        Deep-OC-SORT & 26.0 & 43.6 & 29.5 & 29.5 & 45.3 & 31.7 \\
        BoT-SORT & 24.7 & 32.4 & 28.3 & 25.0 & 35.9 & 33.8 \\\hline
        LOI & 24.7 & 33.1 & 37.4 & 26.9 & 35.2 & 39.9\\
        DRNet & {12.3} & {24.7} & {12.7} & {14.5} & {25.5} & {13.3} \\\hline
    \end{tabular}
    }
    \caption{Performance of existing VIC methods trained with different trajectory labels. 
    }
    \vspace{-2mm}
    \label{tab:weak}
\end{table}

\noindent \textbf{Performs as a Pseudo Trajectory Generator.}
The latent variable $\Omega$ is obtained by solving a problem defined by \cref{eq:f}. The Hungarian Algorithm generates a one-to-one match using $\Omega$ as the cost matrix. 
Based on the matching, we generate pseudo trajectory labels, which are used to train existing VIC methods. We show the results on SenseCrowd in \cref{tab:weak}.
With the generated pseudo label, the performance of all the methods degrades slightly than using the ground truth trajectory labels. This further demonstrates that exact individual-level matching annotations may not be necessary for VIC.

\begin{table}[!tb]
    \centering
    \small
    \setlength{\tabcolsep}{2pt} 
    \begin{tabular}{c ccccccc}
    \hline
         $(n_i,n_{i+\delta})$   &(9,8)&(60,53)&(116,115)&(289,292)&(645,724)\\\hline
         Association~\cite{DRVIC}    &23.8 s& 104.5 s    &174.4 s   &586.9 s  &1663.1 s&\\
         In-Out     &4.4 s & 42.6 s     &53.7 s    &95.0 s&571.9 s \\
    \hline
    \end{tabular}
    \caption{Time cost for a pair of frames with two types of labeling methods. $n_i$ denotes the number of pedestrians in frame $i$.}
    \label{tab:anno}
     \vspace{-7mm}
\end{table}

\noindent \textbf{Label time cost.} 
We compare the time cost of annotating our weakly supervised label and association label~\cite{DRVIC} on five frame pairs with various densities in \cref{tab:anno}.
The annotation time for our weakly supervised label is only about 30\% of that of the association label, making it easier to build a larger dataset for this VIC.
\section{Conclusions}
We propose a new task, Weakly supervised Video Individual Counting~(WVIC), in this paper. Unlike conventional video individual counting tasks, which require expensive trajectory labels, our WVIC task just requires two types of easily available identity-agnostic annotations. This largely reduces the annotation cost and facilitates future works to upscale data size.
Furthermore, we propose a strong baseline for the WVIC task. The key idea is a Group-level Matching Loss~(GML), 
which makes it easier to distinguish inflow individuals from all objects. 
Last, we have extended two existing datasets and built a totally new dataset for the WVIC task.
Our method achieves favorable performance on all three datasets, even compared to supervised methods.
{
    \small    \bibliographystyle{ieee_fullname}
    \bibliography{main}
}


\end{document}